\documentclass[preprint,a4paper]{elsarticle}

\makeatletter
\def\ps@pprintTitle{%
	\let\@oddhead\@empty
	\let\@evenhead\@empty
	\let\@oddfoot\@empty
	\let\@evenfoot\@oddfoot
}
\makeatother

\usepackage{amsfonts,amsmath,amsthm,amstext,amssymb}
\usepackage{graphicx,color}
\usepackage{enumerate}

\usepackage{natbib}

\usepackage{multirow} 
\usepackage{rotating}

\usepackage{marginnote}
\allowdisplaybreaks

\usepackage{lineno}
\newcommand{\jp}[1]{\textcolor{blue}{#1}}

\begin{document}
	\begin{frontmatter}

		\underline{\textbf{Preprint version submitted to Elsevier \hfill July 15, 2024}}\\
		
		\begin{center}
			{\textbf{\Large A reinforcement learning agent for maintenance of deteriorating systems with increasingly imperfect repairs}}
		\end{center}	
		\begin{center}
			Alberto Pliego Marug\'an, Jes\'us Mar\'ia Pinar-P\'erez, and Fausto Pedro Garc\'ia M\'arquez
		\end{center}
		\vspace{5mm}
		\begin{center}
			{\large Published in \textbf{Reliability Engineering \& System Safety - RESS} (ELSEVIER), December 2024.\\}
		\end{center}
		\vspace{5mm}
		\textbf{Cite as:} Marug\'an, A. P., Pinar-P\'erez, J. M., and Garc\'ia M\'arquez, F. P. (2024). A reinforcement learning agent for maintenance of deteriorating systems with increasingly imperfect repairs. Reliability Engineering \& System Safety, 252, 110466.\\
		\begin{center}
			\textbf{DOI:} https://doi.org/10.1016/j.ress.2024.110466
		\end{center}
		\vspace{5mm}
		\begin{flushleft}
			Article available under the terms of the \textbf{CC-BY-NC-ND} licence
		\end{flushleft}
			\vspace{10mm}
			\begin{center}
		The work reported herewith has been financially supported by the Spanish Ministerio de Ciencia, Innovación y Universidades , under Research Grant \textbf{FOWFAM} project with reference: PID2022-140477OA-I00.
	\end{center}
			\begin{figure}[!h]
			\centering
			\includegraphics[trim={0 0 0 0}, width=5cm]{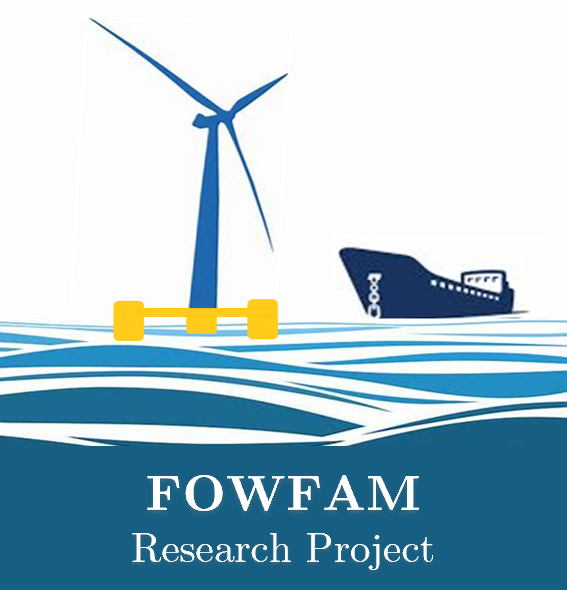}
			\end{figure}

		\vfill
		\textbf{Preprint version submitted to Elsevier \hfill July 15, 2024}
				
		\title{A reinforcement learning agent for maintenance of deteriorating systems with increasingly imperfect repairs} 
			
		\author[add1]{Alberto Pliego Marugán\corref{Alberto}}
		\author[add1]{Jesús M. Pinar-Pérez}
		\author[add2]{Fausto Pedro García Márquez}
		
		\cortext[Alberto]{Corresponding author.  \textit{Email address}: alberto.pliego@cunef.edu}
		\address[add1]{CUNEF Universidad,  Leonardo Prieto Castro 2, Madrid, Spain}
		\address[add2]{Ingenium Research Group, Universidad de Castilla La-Mancha, Av. Camilo José Cela, Ciudad Real, Spain}

		\begin{abstract}
			Efficient maintenance has always been essential for the successful application of engineering systems. However, the challenges to be overcome in the implementation of Industry 4.0 necessitate new paradigms of maintenance optimization. Machine learning techniques are becoming increasingly used in engineering and maintenance, with reinforcement learning being one of the most promising. In this paper, we propose a gamma degradation process together with a novel maintenance model in which repairs are increasingly imperfect, i.e., the beneficial effect of system repairs decreases as more repairs are performed, reflecting the degradational behavior of real-world systems. To generate maintenance policies for this system, we developed a reinforcement-learning-based agent using a Double Deep Q-Network architecture. This agent presents two important advantages: it works without a predefined preventive threshold, and it can operate in a continuous degradation state space. Our agent learns to behave in different scenarios, showing great flexibility. In addition, we performed an analysis of how changes in the main parameters of the environment affect the maintenance policy proposed by the agent. The proposed approach is demonstrated to be appropriate and to significatively improve long-run cost as compared with other common maintenance strategies.  
		\end{abstract}
		
		
	\end{frontmatter}
\newpage
\section{Introduction}\label{Intro}
	
	Globalization and the ultrahigh competitiveness of current and emerging markets necessitate the ongoing modernization and sophistication of engineering systems. However, the development of increasingly complex multicomponent systems introduces myriad ---and often unprecedented--- potential failure mechanisms, which work alongside normal wear-and-tear-related deterioration. Nevertheless, ongoing reliability in the face of increasing sophistication is of paramount importance if such systems are to benefit the industries, businesses, and commercial ventures for which their use is intended. This makes efficient and cost-effective maintenance management essential. 
	
	Maintenance costs are estimated to constitute between 15\% and 70\% of total production costs \citep{thomas2018costs}, with ongoing processes modernization and automation only serving to increase the importance of maintenance. Accordingly, comprehensive maintenance strategies and methodologies have evolved and/or been developed in every industrial and service sector, as exemplified by the automative, food, energy, and pharmaceutical industries, as well as by social services such as education and healthcare \citep{manzini2010maintenance}.

	Maintenance strategies can be divided into two major categories: corrective maintenance (CM) and preventive maintenance (PM). CM is reactive, being initiated after a component fails, while the purpose of PM is to prevent such component failures before they occur. PM further encompasses predictive maintenance (PdM) and condition-based maintenance (CBM), which differ in the way maintenance-need is assessed. PdM involves the use of precise formulas in conjunction with the accurate measurement of environmental factors, such as temperature, vibration, and noise, using sensors or inspections, and maintenance-need is assessed based on analysis of these factors. Accordingly, PdM has the ability to forecast forthcoming maintenance events, making it highly accurate and efficient. Conversely, CBM relies solely on real-time measurements, and maintenance actions are executed once a parameter surpasses a predefined threshold. This means that CBM systems engage in maintenance activities only when required. Furthermore, maintenance strategies are often applied in accordance with a policy having a specific set of characteristics, such as age-replacement, failure-limit, random-age-replacement, repair-cost-limit, and periodic-preventive-maintenance policies \citep{wang2002survey}. 
	
	Improving these maintenance strategies is one of the main challenges facing the emergence of “industry 4.0”, a term for the next-generation developments envisaged for modern and future systems, typically encompassing three main directions, as outlined below \citep{lasi2014industry}:
	
	\begin{itemize}
		\item The first direction concerns adaptability to changing conditions, which includes innovation capability, individualization of products, flexibility, and decentralization. In this field, the availability of all the productive resources of a company is essential to ensure adaptive capacity.
		\item The second direction concerns sustainability and ecological activities. Improving the efficiency of a productive processes implies a reduction of energy waste. Moreover, poor maintenance management can cause additional pollution from productive processes, for instance, leakages in natural gas or petroleum production \citep{wollin2020critical}, poor water quality \citep{vanshkar2019upcoming}, or noise pollution by cars \citep{dierkes2002pollution}. 
		\item The third direction concerns the use of technologies for increasing mechanization, automation, digitalization, and networking. These characteristics depend on the use of electronics, information technologies, real-time data, mobile computing, cloud computing, big data, and the internet of things (IoT) \citep{lu2017industry}. 
	\end{itemize}
	 
	The huge amount of data generated and made available by the third developmental direction will facilitate the creation of intelligent maintenance policies via machine learning techniques. Machine learning is a powerful tool for extracting useful information in this massive data environment. Current literature contains numerous algorithms for data-driven decision making in the field of maintenance, and research interest in machine learning for maintenance management is clearly increasing. This interest is strengthened by the necessity of data processing and the increasing importance of the maintenance of systems. 

	This paper is centered in one of the three major paradigms of machine learning: reinforcement learning (RL). RL seeks a set of optimal actions by an agent within a defined environment for maximizing rewards. With RL, the final reward is cumulative, since it is the result of progressive actions corresponding to a specific action policy. Accordingly, RL shows enormous promise for addressing computational problems in a way that achieves long-term goals \citep{sutton2018reinforcement}. 
	
	Clearly, the use of machine learning techniques has significantly increased in recent years, but the increase in the use of RL is even more significant. \jp{It should be noted that today} the number of publications mentioning RL in the field of maintenance is almost 20-times greater than \jp{a decade ago}. 
	
	The objective of this study \jp{is} to explore the capacity of RL agents to generate policies that improve the maintenance of deteriorating systems. Any improvement in maintenance policy will be assessed in terms of long-term costs. \jp{The proposed model can be applied in  industrial systems or components subjected to deterioration. For instance, maintenance of renewable energy systems such as wind turbines or solar panels, maintenance of elevators in commercial buildings, conveyor belts in warehouses, irrigation systems in agriculture, office equipment such as printers or HVAC systems, public lighting systems, etc.  It must be mentioned that our RL agent has been developed to minimize long run cost rates, i.e. to improve maintenance from a purely economic perspective. Hence, as the deterioration may increase at intolerable levels, this methodology is not applicable in its current formulation to critical safety systems such as maintenance of aircrafts, nuclear plants, etc, where failures can be catastrophic.}
	
	The main novelty of this study lies in the combination of a maintenance model in which each repair is less effective as more repairs are conducted, and a RL agent whose structure directly addresses the maintenance problem without the need to discretize the degradation state. This combination significantly aligns the model with reality, where the degradation process is continuous, repairs are imperfect, and systems are affected by consecutive repairs.
	
	The remaining content of this paper is structured as follows: Section \ref{S2} reviews the most pertinent literature on deteriorating systems and maintenance models. Similar studies are presented to highlight the main contributions of our work. Section \ref{S3} briefly explains the main concepts of RL and the Double Deep Q-Network (DDQN) structure employed in this work. Section \ref{S4} presents the proposed system to be subjected to degradation and the possible maintenance actions. Section \ref{S5} describes the environment and the RL agent proposed in this paper. Section \ref{S6} shows different scenarios to be analyzed, the main results, and a comparison of the proposed maintenance policy with other conventional policies. Finally, section \ref{S7} presents the main conclusions of our work.
	
\section{Stochastic degradation processes and RL maintenance}\label{S2}
	Most systems employed in production processes are subject to degradation. A deteriorating system can be defined as a system with an increasing probability of the occurrence of failures \citep{kaminskiy2010gini}, i.e., a decreasing reliability over time. However, most of these systems can be maintained or repaired. Constructing accurate models that define degradation processes is essential for operations and maintenance purposes and product design. Such models provide valuable information of the reliability, \jp{remaining useful life (RUL)}, and actual conditional state of a product during its lifecycle. 
	
	An interesting classification of the main degradation models was proposed by Kang et al. \citep{rui2020model}. In terms of this classification regime, this paper is focused on monotonical stochastic degradation processes (SDPs) with single-mechanism degradation. The term "monotonical" indicates that the degradation is irreversible, i.e., the state of the system worsens over time unless a maintenance activity is carried out. This situation corresponds to most actual degradation phenomena. According to Peng and Tseng \citep{peng2009mis}, a good stochastic model should satisfy three main properties: clear physical explanation; easy formulation; and adaptability to exogenous events. Whitin this field, the most common stochastic processes satisfying these properties are gamma, inverse Gaussian, and Wiener processes for continuous degradation, and Markov chains for discrete degradation modelling.
	
	In this paper, we propose a continuous monotonic degradation model based on the gamma stochastic process. Gamma-process-based models were introduced in 1975 by Abdel-Hameed \citep{abdel1975gamma} and have since been widely used to model deterioration. An extensive review of gamma degradation processes is provided by van Noortwijk \citep{van2009survey}. 
	
	The increasing importance of maintenance has led to the development of policies and algorithms to obtain optimal maintenance policies \citep{pliego2016optimal} considering SDP. However, it is not possible to define an optimal maintenance for all systems since their maintenance does not always have the same goals and must be adapted to each type of system. There are numerous reported methodologies for the maintenance of systems subject to SDP, including value iteration algorithms \citep{cheng2023maintenance}, stochastic filtering \citep{zhang2015degradation}, multi-objective optimization \citep{shahraki2020selective}, stochastic programming formulation \citep{leo2022condition,ruiz2020multi}, and others \citep{khatab2014availability,chuang2020condition,wang2021joint}. In addition to these algorithms and methods, some researchers have recently employed the capacities of RL to improve different aspects of maintenance management. Some RL-based approaches are employed to aid the maintenance tasks on safety-critical systems, i.e., those systems whose failure or fault entails catastrophic consequences \citep{bowen1993safety}. Therefore, the main objective in the maintenance of these types of system is to maximize the system’s reliability. For instance, Aissani et al. \citep{aissani2009dynamic} developed a multi-agent approach for effective maintenance scheduling in a petroleum refinery. They achieved a continuous improvement of solution quality by employing a SARSA algorithm. Mattila and Virtalen \citep{mattila2011scheduling} proposed two formulations for scheduling the maintenance of fighter aircraft via RL techniques, i.e., $\lambda$-SMART and SARSA algorithms, and achieved improved results with respect to heuristic baseline policies. However, RL algorithms are mostly employed in non-safety-critical systems where the main goal of maintenance is to maximize profit, which does not always coincide with maximizing reliability. In this field, RL has been employed for several system types, including manufacturing and production systems used in flow line manufacturing \citep{wang2016multi}; civil infrastructure systems used for bridges \citep{wei2020optimal}, pavements \citep{yao2020deep}, and roads \citep{tanimoto2021combinatorial}; transportation systems used in the maintenance of ships \citep{le2021reinforcement}; power and energy systems used in offshore wind farms \citep{chatterjee2020deep}, power grids \citep{rocchetta2019reinforcement,yang2021optimization}, and energy storage systems \citep{wu2021intelligent}; and other more specific systems such as those used in medical equipment \citep{ma2021research} and Mobile Edge Computing systems \citep{wang2019smart}. An exhaustive review of the use of RL for maintenance of different types of systems is provided by Marugán \citep{marugan2023applications}. 
	
	In this paper, we are mainly interested in RL-based models for deteriorating systems. Several approaches can be found in this field, for instance, Andriotis and Papakonstantinou \citep{andriotis2021deep} proposed a stochastic optimal control framework for the maintenance of deteriorating systems with incomplete information. They considered stochastic, non-stationary, and partially observable ten-component deteriorating systems in four possible degradation states. They employed a DDMAC structure, which was compared with several baseline maintenance polices, such as fail replacement (FR), age-periodic maintenance (APM), age-periodic inspections with CBM (API-CBM), time-periodic inspections with CMB (TPI-CBM), and risk-based inspections with CBM (RBI-CBM). Their proposed agent clearly outperformed all the baselines. Peng and Feng \citep{peng2021reinforcement} introduced a study addressing the decision-making problem of CBM for lithium-ion batteries, representing their capacity degradation with a Wiener process. To tackle this problem, they employed an algorithm known as Gaussian process with reinforcement learning (GPRL). Unlike the prevailing approaches, which primarily focus on maximizing discounted rewards, the GPRL algorithm aims to minimize long-term average costs. This alternative approach demonstrated superior performance in comparison with the conventional methodology. Wang et al. \citep{wang2021integrated} employed a Q-Learning-based solution in a multi-state single machine with deteriorating effects. They developed a PM strategy that combined time-based PM and CBM, and they employed a discrete deterioration model using a Markov chain with four possible states, which was used to demonstrate the high performance and flexibility of the proposed RL approach. Zhang et al. \citep{zhang2021model} proposed a customized Q-Learning method called Dyna-Q to deal with a system with a large number of degradation levels and where the degradation formula is unknown. Due to the number of possible states, this model can be considered halfway between a discrete and continuous degradation model. Adsule et al. \citep{adsule2020reinforcement} studied degradation in terms of the wear of a component. They considered a Gaussian model for the stochastic degradation, and a SMART RL algorithm was employed. The agent was able to obtain an optimal or near optimal policy to determine maintenance actions and inspection scheduling. Zhao and Smidts \citep{zhao2022reinforcement} proposed a case study of a pump system used in nuclear power plants with a Gamma deterioration process. The problem was presented as a partially observable Markov decision problem where knowledge of the system is improved with Bayesian inference. Zhang et al. \citep{zhang2020deep} modelled the SDP for a multi-component system based on the compound Poisson and gamma processes. They employed a DQN algorithm to optimize the CBM policy under different scenarios. The gamma process is also employed by Yousefi et al. \citep{yousefi2020reinforcement} who proposed a Q-Learning algorithm to find policies in a repairable multi-component system being subjected to two failure processes - degradation and random shocks. Despite considering a continuous SDP, they discretized the deterioration into four levels, allowing them to describe a discrete MDP. 
	
	Compared with these previous studies, the main contributions of this study are:
	\begin{itemize}
		\item A deteriorating system and maintenance model that consider imperfect maintenance with an important novelty with respect to the literature found. The maintenance model considers imperfect maintenance, and repairs become increasingly imperfect as more repairs are undertaken. This behavior is represented by a truncated normal distribution whose mean depends on the number of previous repairments done over the system.
		\item	The implementation of a RL agent, which allows for the improvement of maintenance policies compared to conventional maintenance strategies. The proposed methodology allows generation of maintenance policies without the need for setting preventive maintenance thresholds. 
		\item	A study of the RL agent performance in different scenarios and a numerical analysis of the effect of changing key parameters (costs of maintenance activities, inspection intervals, degradation rate) on the maintenance policies generated by the agent. This article aims to demonstrate not only that RL techniques are suitable for generating maintenance policies in deteriorating systems, but also that they can be extremely flexible facing parameter changes.
		\item	Our RL agent can operate in a continuous deterioration space without the need for a discretization process. 
	\end{itemize}
		
\section{RL framework}\label{S3}
\begin{figure}[!h]
	\centering
	\includegraphics[trim={0 0 0 0}, width=10cm]{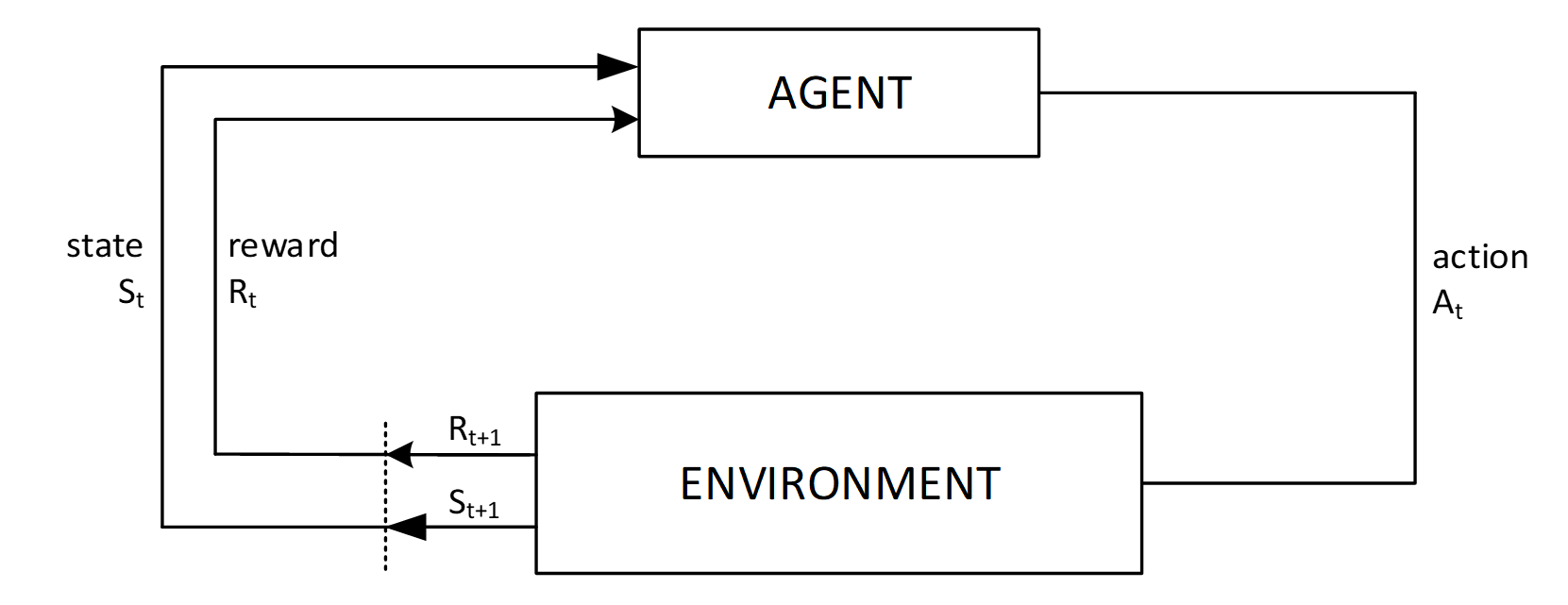}
	\caption{General RL structure. Adapted from \citep{sutton2018reinforcement}}\label{fig:RLstruc}
	\end{figure}

	RL is a computational strategy that proposes an iterative trial-and-error interaction between an agent and its environment. This process leads the agent to generate a maintenance policy aimed at maximizing a specific reward. Key components of an RL system encompass the agent, the available actions, the associated rewards, and the environmental context. The interaction between the agent and environment is often depicted as illustrated in \jp{Figure \ref{fig:RLstruc}} .
	
	Interaction between agent and environment is typically explained within the formal framework of Markov decision processes (MDPs) \citep{sutton2018reinforcement}. \jp{A MDP} problem is formed by the pertinent tuple $\left( \mathcal{S}, \mathcal{A}, \mathcal{T}, \mathcal{R} \right)$, where $\mathcal{S}$ denotes \jp{the state space}, $\mathcal{A}$ stands for \jp{the action space}, $\mathcal{T} \colon \mathcal{S} \times \mathcal{A} \times \mathcal{S} \rightarrow [0,1]$ is the transition probability function providing the probability of transitioning from state $s$ to $s'$ due to action $a$, and $\mathcal{R} \colon \mathcal{S} \times \mathcal{A} \rightarrow [0,1] $ stands as the reward function, stipulating the reward due to a transition from state $s$ to $s'$ \citep{sutton2018reinforcement}. 
	
	\jp{In reinforcement learning, the agent's objective is defined by a special signal known as the reward, which is transmitted from the environment to the agent. At each time step, the reward is a single numerical value, denoted as $r_t \in \mathbb{R}$. The sequence of rewards after the time step $t$ is $r_{t+1}, r_{t+2}, r_{t+3},\ldots$. The cumulative reward ($G_t$) represents the discounted reward or the sum of future rewards from the time $t$.  For a trajectory of finite length $K$ within the environment, $G_t$ is defined by equation (\ref{eq:GT})}.
	\begin{align}\label{eq:GT}
		G_t=\sum_{k=0}^{K} \gamma^k r_{t+k}
	\end{align}
	\jp{where $\gamma \in [0,1]$ is a discount factor that determines the relevance of the future rewards and forces the convergence for infinite-horizon returns. $k \in [0,K]$ is a subindex, being $K$ the total number of future recompenses that the agent will receive until the end of the current episode.} Rewards are used by the agent to generate a policy $\pi \colon \mathcal{S} \times \mathcal{A} \rightarrow [0,1]$, i.e., a function providing the probability distribution of each action $ a \in \mathcal{A} $ and each possible state $s \in \mathcal{S}$. Following a given policy $\pi$, a value function and an action-value function can be defined as:
	\begin{align}\label{eq:Vpi}
		V^\pi(s)= \mathbb{E}_\pi \left[ \sum_{k=0}^{K} \gamma^k r_{t+k} \mid s_t = s \right] 
	\end{align}
	\begin{align}\label{eq:Qpi}
		Q^{\pi} (s,a) = \mathbb{E}_\pi \left[ \sum_{k=0}^{K} \gamma^k r_{t+k} \mid s_t = s , a_t = a \right] 
	\end{align}
	
	\jp{The policy $\pi$ maps states to the probability of selecting each possible action. Therefore, if the agent follows the policy $\pi$ at a certain time $t$, then $\pi\left(a\middle| s\right)$ represents the probability of choosing the action $a$ given a state $s$. Therefore, this policy depends only on the current state and not on the sequence of states and actions that preceded it, being aligned with the principles of MDP.}
	
	The main goal \jp{of the RL agent is} to find the policy $\pi^\ast$ that maximizes the expected reward, satisfying the Bellman optimality equations (\ref{eq:Qast}) and (\ref{eq:Vast}).
	\begin{align}\label{eq:Qast}
		Q^{\ast} (s,a) = \sum_{s',r} p\left( s',r \mid s, a \right) \left[ r + \gamma \max_{a'} Q^{\ast} (s',a') \right]  
	\end{align}
	
	\begin{align}\label{eq:Vast}
		V^\ast (s)=\max_{a \in \mathcal{A}(s)} 	Q^{\pi^\ast} (s,a) = \sum_{s',r} p\left( s',r \mid s, a \right) \left[ r + \gamma V^\ast (s') \right]  
	\end{align}
	
	Therefore, $\pi^\ast$ being the policy that maximizes the value functions, equations (\ref{eq:Piast}) and (\ref{eq:Piast2}) will provide the optimal policy:
	\begin{align}\label{eq:Piast}
		\pi^\ast = \arg \max_\pi V^\pi (s)  
	\end{align}
	\begin{align}\label{eq:Piast2}
		\pi^\ast = \arg \max_\pi Q^\pi (s,a)  
	\end{align}
	
	These optimal policies can be attained by following different strategies. Depending on the characteristics of environment, different algorithms can be employed. A review of RL algorithms can be found in Shakya et al. \citep{shakya2023reinforcement}. 
	
	In this paper, we employ the DDQN algorithm, proposed originally by Hasselt \citep{hasselt2010double}. This algorithm, which is derived from the Deep Q-Network (DQN) algorithm, addresses the problem of Q-value overestimation, which is frequently provided by the standard DQN algorithm proposed by Mnih et al \citep{mnih2015human}. A DQN consists of a neural network that, given a state $s$, produces a vector of action values $Q(s;\theta)$, where $\theta$ represents the parameters of the neural network. The DQN algorithm incorporates three essential components: first, a neural network (main neural network) with parameters $\theta$, which is employed to estimate Q-values of the current state $s$ and $a$; a second neural network (target neural network) with parameters $\theta'$ used to approximate the Q-values of the next state $s’$ and next action $a'$; a replay memory used to store the experiences for the learning process and the implementation of a target network with parameters $\theta$ \citep{mo2019decision}. The Bellman equation for a DQN is: 
	\begin{align}\label{eq:Bell}
		Q (s,a;\theta) =  r + \gamma Q \left( s', \max_{a'} Q \left( s', a';\theta' \right) \right) 
	\end{align}
	
\begin{figure}[!h]
		\centering
		\includegraphics[trim={0 0 0 0}, width=10cm]{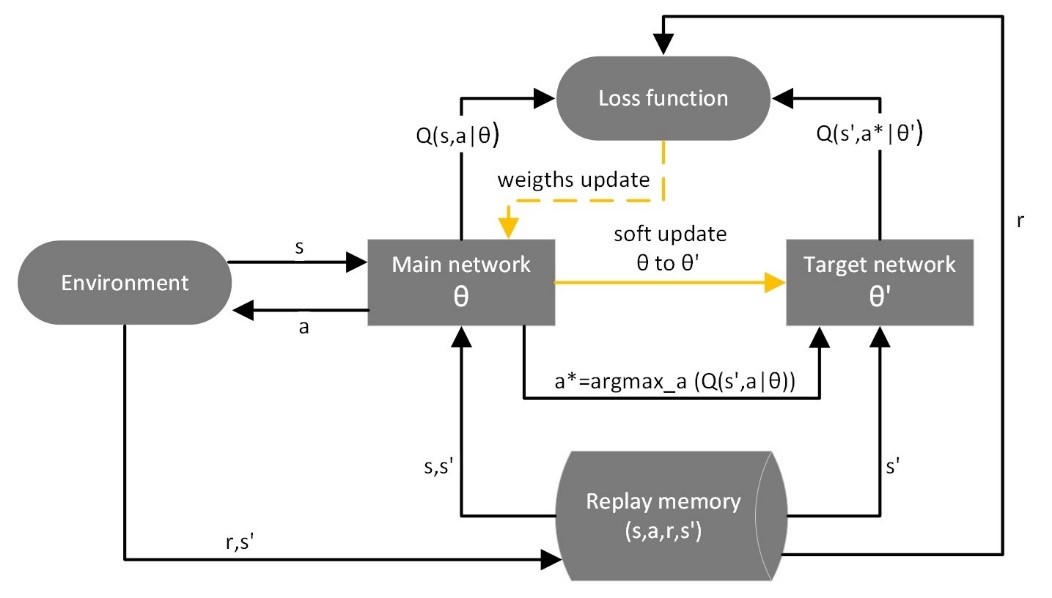}
		\caption{Double Deep Q-Network architecture}\label{fig:DDQN}
	\end{figure}

	The main difference between a DDQN and a DQN is that the process of action selection and action evaluation are separate in a DDQN, as the target Q-values are determined by actions selected by the main network, while their Q-values are estimated using the target network. This adjustment effectively eliminates overestimation bias, leading to more precise Q-value estimates and enhanced training stability. Considering these changes, the Bellman equation for a DDQN results in: 
	\begin{align}\label{eq:Bellddqn}
		Q (s,a;\theta) =  r + \gamma Q \left( s', \arg \max_{a'} Q \left( s', a';\theta \right);\theta' \right) 
	\end{align}
	The main goal of DQNs and DDQNs is to estimate Q-values through deep neural networks, which is especially useful when the state space is too large to be collected in a table (as a Q-learning algorithm does). The architecture of a DDQN algorithm is illustrated in \jp{Figure \ref{fig:DDQN}} .
	
	The decision to use a DDQN in this study was not arbitrary since it has been demonstrated that DDQN agents outperform other algorithms when dealing with very large state spaces. In this paper, we do not discretize the degradation level, so the state space is continuous while the action space is discrete. This features make DDQNs highly suited to work in this environment, as demonstrated in other studies \citep{li2022learning,raghu2017continuous,zhang2022ddqn}. Other suitable architectures, such as proximal policy optimization (PPO) and trust region policy optimization (TRPO) have been assessed for our environment, but they provided inferior results. 

\section{Proposed degradation process and maintenance model}\label{S4}
	This paper proposes a new approach to optimize the CBM policy for a gradually deteriorating single-unit system subjected to SDP. Degradation is modelled by a homogeneous gamma process. The proposed model is based on Marugan et al. \citep{marugan2023comparative}.

	The gamma process, which is assumed to be strictly increasing over time if no maintenance action is carried out, can be formulated as $(X_t)_{t\geq0}$. Let the random variable $X_t$ stand for the deterioration state of the system at time $t$, where $X_0 = 0$ and $t\geq0$. The degradation increment $\Delta X(t,\Delta t) = X_{t+\Delta t}-X_t$ is a continuous random variable following a gamma distribution with shape parameter $v(t,\Delta t)$  and scale rate $\beta$. Therefore, $\Delta X \sim \varGamma \left( v(t,\Delta t),\beta\right)$ and its probability density function (pdf) is: 
	\begin{align}
		f \left( t,\Delta t, x \right)= Pr \left(\Delta X = x \right) = \frac{x^{v(t,\Delta t)-1}}{\varGamma \left( v(t,\Delta t) \right)} \beta^{v(t,\Delta t)} e^{-\beta x}, \quad \forall \hspace{0.1cm} x\geq 0
	\end{align}
	
	If $v(t,\Delta t)$ is a linear function, the model results in a stationary gamma process; otherwise, the process becomes non-stationary. 
	
	The cumulative density function is: 
	\begin{align}
		F \left( t,\Delta t, x \right)=\frac{\gamma 	\left( v(t,\Delta t) , \left( \beta x \right) \right) }{\varGamma \left( v(t,\Delta t) \right)} 
	\end{align}
	where $\gamma(\cdot)$ is the lower incomplete gamma function. 
	
	The survival function can be defined by:
	\begin{align}
		\bar{F} \left( t,\Delta t, x \right)=1-F \left( t,\Delta t, x \right)=\frac{\varGamma \left( v(t,\Delta t) , \left( \beta x \right) \right) }{\varGamma \left( v(t,\Delta t) \right)} 
	\end{align}
		
	Besides the deterioration model employed to describe stochastically the state of the system, it is essential to define the way such states are obtained. In this field, continuous monitoring, which provides the system condition in real time, is the most accurate method. Continuous monitoring allows anomalies to be detected at initial stages, allowing maintenance actions to be performed immediately \citep{hao2020condition}. However, factors such as costs, technological limitations, legal issues, or other limitations make continuous monitoring inadequate for some systems. In such cases where continuous monitoring is not suitable, the deterioration state is often obtained via planned inspections. In this paper, we propose planned inspections to determine the state of the system. We consider perfect inspection, i.e., the system state is revealed with certainty. Additionally, we assume that these inspections are instantaneous, so that the duration of the inspection is negligible. Inspections are executed at times $\left(T_n\right)_{n \in \mathbb{N}}$ with $(T_0 )=0$. Let $T_n^-$ and $X_{T_n^-}$ be the time and the state of the system just before the inspection at time $T_n$, respectively. 
	
	Regarding the maintenance characteristics, we consider an imperfect maintenance for a repairable unit system. Two types of maintenance activities have been considered in this work: replacements and repairs. Like inspections, the maintenance interventions are assumed to be instantaneous, and the effect of these actions is observable immediately after the inspection, i.e., at time $T_n^+$. The available maintenance actions are similar to the CBM model presented by Zhang et al. \citep{zhang2021model} but the behavior of the model presented herein is totally different. These maintenance activities are:
	\begin{description}
		\item [\textit{Replacements} $\left( R \right)$:] A replacement leads the system to an "as good as new" (AGAN) state, i.e., the deterioration of the system after any replacement is $X_{T_n}^R = 0$. If this action is performed when the deterioration is above a failure threshold $L$, i.e., the deterioration reaches an unacceptable value above which the system will fail, the action is said to be a \textit{corrective replacement}. However, if the action is carried out below the threshold, then the action is a \textit{preventive replacement}, and no downtime costs are computed.
		\item [\textit{Repairs} $\left( P \right)$:] This maintenance task is assumed to be imperfect. Several previous studies have modeled imperfect maintenance. We combine some characteristics from the models proposed in Huynh \citep{huynh2019hybrid} and van Bérenguer \citep{van2012condition} to determine the effect of an imperfect repair. We model the effect of imperfect repairs by subtracting a certain amount from the current deterioration level. This amount is sampled from a random distribution with a memory effect, i.e., it depends on the previous repairs. This memory is represented by assuming that, after a repair, the system cannot return to a deterioration state lower than that reached in the previous maintenance action. Note that we do not consider a corrective repairment; we assume that when the system has failed, it is necessary to reset the degradation to 0.
	\end{description}	
		
	Let $X_{T_n}^P$  be the degradation state after a preventive repair action. We consider that $X_{T_n}^P = X_{T_n^-} - Z_n$, where $Z_n$, called the maintenance gain, is a continuous random variable distributed as a truncated normal distribution whose density is :		
	\begin{align}
		g_{\mu , \sigma , X_{T_n^-}} (x) = \frac{1}{\sigma} \frac{\phi \left( \frac{x-\mu}{\sigma} \right)}{\Phi \left( \frac{X_{T_ {\bar n}}-\mu}{\sigma}\right) -  \Phi \left( \frac{X^M -\mu}{\sigma}\right) } I_{ \left[ X^M , X_{T_n^-}\right]} \left(x\right)
	\end{align}	
	where:
	\begin{itemize}
		\item $\phi (\cdot)$ and $\Phi (\cdot)$ are the probability density and cumulative distribution function of the standard normal distribution respectively;
		\item $X^M$ is the deterioration value after the immediate previous maintenance activity;
		\item $I_{\left[X^M,X_{T_n^-} \right]}  (x) = 1$ if $X^M \leq x \leq X_{T_n^-}$ and $I_{\left[X^M,X_{T_n^-} \right]}  (x) = 0$, otherwise;
		\item $\mu$ and $\sigma$ are the mean and standard deviation of the truncated normal distribution. 
	\end{itemize}

	\jp{Similarly to} \citep{do2015proactive}, we assume that $\mu=\frac{X^M + X_{T_n^-}}{2}$ and $\sigma=\frac{X^M + X_{T_n^-}}{6}$.	\\
	
	\jp{The use of a truncated normal was proposed originally in reference \citep{van2012condition} to model the maintenance gain. This model is appropriate because it captures the variability of the system deterioration after an imperfect repair and allows for considering practical limits in the model. In our approach, the system deterioration after a repair action is bounded by the current deterioration state and the deterioration state after the previous maintenance intervention. }	
	
\begin{figure}[!h]
	\centering
	\includegraphics[trim={0 1cm 0 0}, width=11cm]{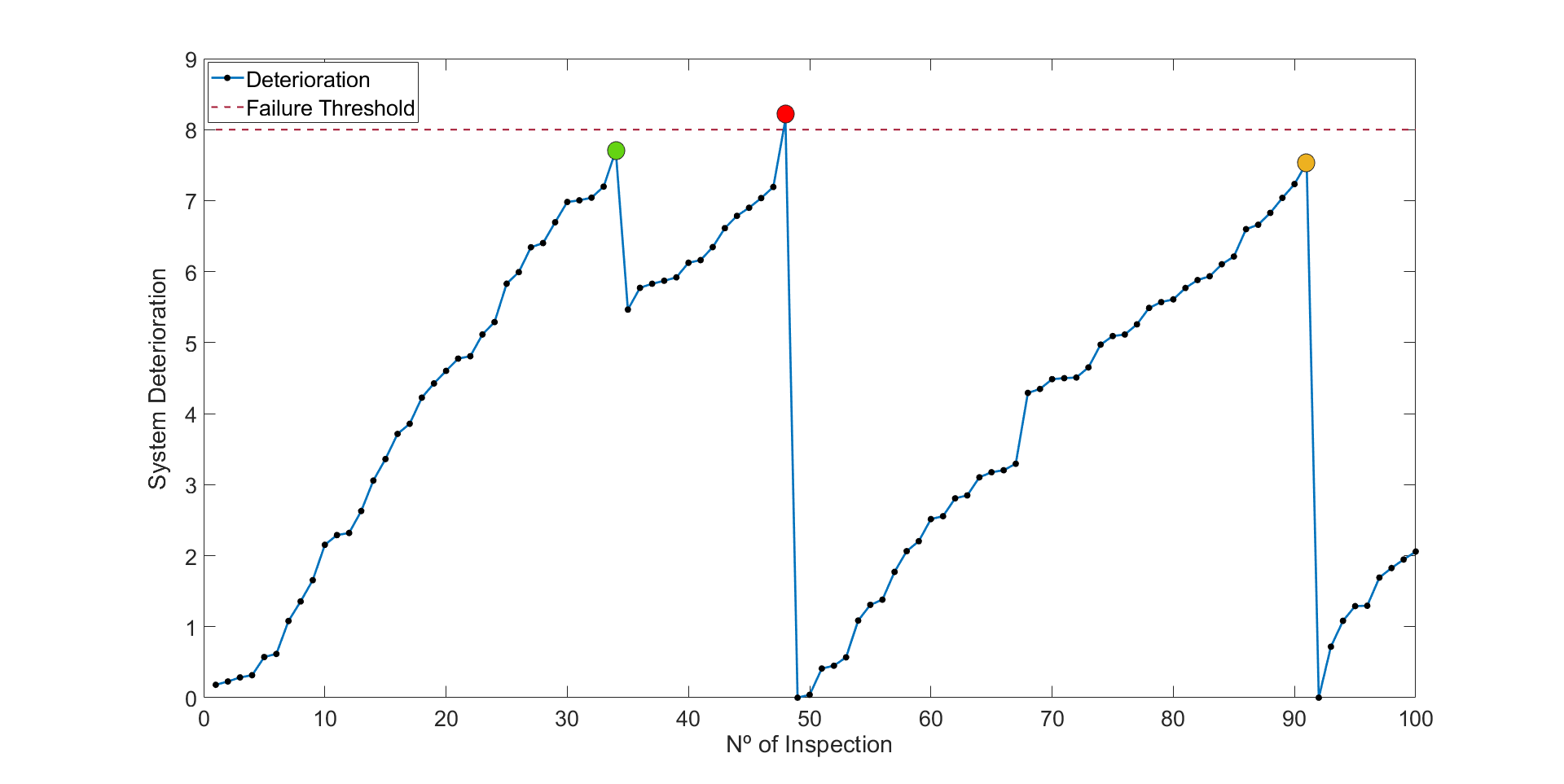}
	\caption{An example of the degradation process. (green circle: preventive repair; red circle: corrective replacement; orange circle: preventive replacement)}\label{fig:EoDP}
	\end{figure}

	An illustrative example of the proposed model is shown in \jp{Figure \ref{fig:EoDP}}, which shows the increasing deterioration and the maintenance actions allowed by the model. 

	\jp{Note that we consider the working state of system to be binary: it is either functioning or not. The deterioration does not affect the performance of the system unless the failure thresholds is surpassed.} Note that most literature on CBM consider a preventive maintenance threshold. One of the advantages of our approach is that this preventive threshold is not necessary since the RL agent will determine the best moment to perform either a corrective or a preventive maintenance. However, in our model, we define a corrective threshold $L$ to determine a system failure. 
	
\section{Description of the environment and the RL Agent}\label{S5}

\subsection{State space representation}
	The RL agent needs to be input with the current state of the environment. This state will not only depend on the current deterioration of the system, but also on the number of preventive repair actions performed since the last reset to an AGAN state. Additionally, as deterioration is a time-dependent process, time is essential to calculate the next state. \jp{The time period between two consecutive inspections $(\Delta t)$ is defined to calculate the increment of deterioration between the state $S^{T_n}$ and the state  $S^{T_{n+1}}$. The definition of this parameter does not affect the Markovian properties of the process, since the value of $\Delta t$ is constant and does not depend on any past event.} The RL agent only needs to act for discrete times, i.e., an action is only ordered after a certain inspection; so the system state turns in a continuous degradation and discrete time state space. The degradation of the system at the inspection n is represented by $X_{T_n}$. As well as the system degradation, it is also necessary to input the degradation after the previous maintenance action, represented by $X^M$. 
	
	Therefore, this system state space will be given by:
	\begin{align}
		S^{T_n}=\left\lbrace X_{T_n}, X^M \right\rbrace \quad \text{with} \quad X_{T_n} \geq 0 \quad \text{and} \quad 0 < X^M \le  X_{T_n}
	\end{align}
	
\subsection{Action space representation}
	According to the model description in Section \ref{S4}, the action space for each inspection $n$ is $\mathcal{A} = \{ a_0,a_1,a_2 \}$, where:
	\begin{itemize}
		\item $a_0$ corresponds to "\textit{no maintenance action}" after the inspection $n$. System deterioration will continue according to the SDP defined in Section \ref{S4}. After action $a_0$  at time $T_n$, the system state is $S^{T_n} = \{X_{T_n},X^M\}$ with $X_{T_n} = X_{T_n}^-$.
		\item 	$a_1$ is a "\textit{preventive repair action}" which leads the system deterioration to any state between $X_{T_n}^-$ and $X^M$ according to the truncated normal model presented in Section \ref{S4}. After action $a_1$  at time $T_n$, the system state is $S^{T_n} = \{X_{T_n},X_{T_n}\}$ with $X_{T_n}\leq X_{T_n}^-$.
		\item $a_2$ refers to a "\textit{replacement action}". Note that action $a_2$ encompasses both preventive and corrective replacements. Being preventive or corrective only depends on the state of the system when the action is performed. This action will provide different rewards regarding the state of the system; however, the consequence for the system state after the action is identical. Both actions set the system to an AGAN state. After action $a_2$  at time $T_n$, the system state is $S^{T_n}=\{0,0\}$. 
	\end{itemize}

\subsection{Rewards definition}
	The main purpose of this paper is to improve the maintenance strategy from an economic perspective. This objective is to minimize maintenance long-run cost. As aforementioned, the RL agent is created to maximize a long-term reward. These rewards will be defined in the function of both the deterioration state and the action selected by the agent. 

	Let $C_P$  and $C_R$ stand for the costs of preventive repair and replacement actions, respectively. As mentioned before, the inspections are assumed to be instantaneous, but if the system fails, we consider that for the time between consecutive inspections, the system is not functioning and therefore there is a loss of production due to the downtime, represented by $C_{down}$. Then, the reward at time $T_n$ is defined as:
	\begin{equation} 
	\jp{r}_{T_n} (a_{T_n},X_{T_n}^-) =\left\{
	\begin{array}{rl} \label{eq:llave2}
	0 & \text{for} \quad a_{T_n} =a_0  \quad \text{and} \quad X_{T_n}^- < L\\
	-C_p & \text{for} \quad a_{T_n} =a_1 \quad \text{and} \quad X_{T_n}^- < L \\
	-C_R & \text{for} \quad a_{T_n} =a_2 \quad \text{and} \quad X_{T_n}^- < L \\
	-C_R - C_{down} & \text{for} \quad a_{T_n} =a_2 \quad \text{and} \quad X_{T_n}^- \geq L 
	\end{array} \right.
	\end{equation}

\subsection{Agent definition and training}	
	The DDQN algorithm was implemented in MATLAB with the following hyperparameters that correspond to the main hyperparameter configuration predefined in MATLAB.
	\begin{itemize}
		\item Exploration options: Epsilon decay $=0.005$; Epsilon max $=1$; Epsilon min $=0.01$; 
		\item Agent options: Sample time $=1$, Discount factor $=0.99$, Batch size $=64$, Experience buffer length $=10000$; 
		\item Optimizer options: Optimizer: ADAM; Learn rate $=0.01$; Gradient decay factor $=0.9$.
	\end{itemize}
	
	Training options have been set as follows: Maximum episodes: $50000$, and Maximum Episode Length $= 500$. The stopping criteria has been set to reach the maximum episodes number.
	
	Note that the agent performance might be improved by tuning these parameters. However, our hyperparameter configuration is sufficient to demonstrate that a DDQN agent is able to generate successful maintenance policies under each scenario and to observe the behavior of the agent when the environment changes. 

\section{Case Studies}\label{S6}

\subsection{Numerical experiments}
	Seven different scenarios to obtain information of the performance of the proposed RL agent in different case studies have been considered. The data values employed in this study are based on the parameters of the case study presented by Zheng et al. \citep{zheng2020optimal}. Case 2* will be considered as the baseline case. It is worth mentioning that, for all cases, $C_R=3500$ since we are interested in the effect of the ratio repair/replacement costs and $v(t)=0.0115t$ since we assume the gamma deterioration process to be homogeneous. The rest of the parameters will vary between cases, as shown in Table \ref{tab:Tab1} . 
	
		\begin{table}[htbp]
		\centering
		\caption{Case studies parameters}
		\footnotesize
		\begin{tabular}{p{1.2cm}p{3.2cm}ccccc}
			\cline{2-7}
			& Description &  $\beta$     & $C_P$  &  $C_{down}$     & $L$      &  $\Delta t$ \\
			\hline
			\textbf{Case 1} & Reduced repair costs & $4.63$  & $\textcolor{red}{300}$ & $2000$  & $8$     & $100 $\\
			\textbf{Case 2*} & Baseline & $4.63$  & $600$   & $2000$  & $8$     & $100$ \\
			\textbf{Case 3} & Increased repair costs & $4.63$  & $\textcolor{red}{1500}$ & $2000$  & $8$     & $100$ \\
			\textbf{Case 4} & Increased failure limit  & $4.63$  & $600 $  & $2000 $ & $\textcolor{red}{12}$ &$100$ \\
			\textbf{Case 5} & Reduced downtimes cost & $4.63 $ & $600$   & $\textcolor{red}{500}$ & $8 $    & $100$ \\
			\textbf{Case 6} & Slower degradation  &$ \textcolor{red}{6.5}$ & $600 $  & $2000$  &$ 8 $    & $100$ \\
			\textbf{Case 7} & Longer inspection period &$ 4.63 $ & $600 $  & $2000 $ &$ 8$     & $\textcolor{red}{150}$ \\
		\end{tabular}%
		\label{tab:Tab1}%
	\end{table}%
	
	Once the different case studies are analyzed, we calculate the long-run cost rate as a performance indicator of the proposed policy. The total costs of maintenance up to moment $t$ can be defined by: 
	\begin{multline}
		C(t) = C_P (t) + C_R (t) + C_{down} (t) = C_P N_P (t) + C_R N_{PR} (t) \\ +  \left( C_R + C_{down} \right) N_{CR} (t)
	\end{multline}	
	where $C_P(t)$, $C_R(t)$, and $C_{down}(t)$ are respectively the cumulative costs of preventive repair, corrective replacement, and production-loss costs due to unavailability of the system. These cumulative costs are given by the fix costs $\left(C_P,C_R,C_{loss}\right)$ and $N_P(t), N_{PR}(t)$, and $N_{CR}(t)$, which are the random number of repairs, preventive replacements, and corrective replacements, respectively, in the period $[0,t)$. 
	
	The  long-run cost rate can be calculated by:
	\begin{align}
		EC_{\infty} = \lim_{t\rightarrow\infty} \left[ \frac{E[C(t)]}{t}\right] = \frac{E[C(S_i)]}{E[S_i]}
	\end{align}
	where $S_i$ refers to the $i^{th}$ renewal cycle (see \jp{Figure \ref{fig:EoDP}}). 
	
	It is known that the system has regenerative properties since corrective replacements are forced when the deterioration surpasses the failure threshold, i.e., the deterioration will come to state 0 in a finite time. Due to such regenerative properties, the long-run cost rate can be calculated through the expected values in a single renewal cycle, a renewal cycle being the period from one replacement to the time just before the next replacement. Therefore, the long-run cost rate can be approximated by:
	\begin{align}
		EC_{\infty} = \frac{C_P E[N_P (S_1)] + C_R E[N_{PR} (S_1)] + \left( C_R + C_{down} \right) E[N_{CR} (S_1)]}{E[S_1]}
	\end{align}
	
	The required expected values will be numerically obtained using the Monte Carlo method. Additionally, to be sure that results are statistically significant, the long-run costs rates will be estimated through confidence intervals.
	
\subsection{Results}
\begin{figure}[!h]
		\centering
		\includegraphics[trim={1cm 0 1cm 1cm}, width=10.5cm]{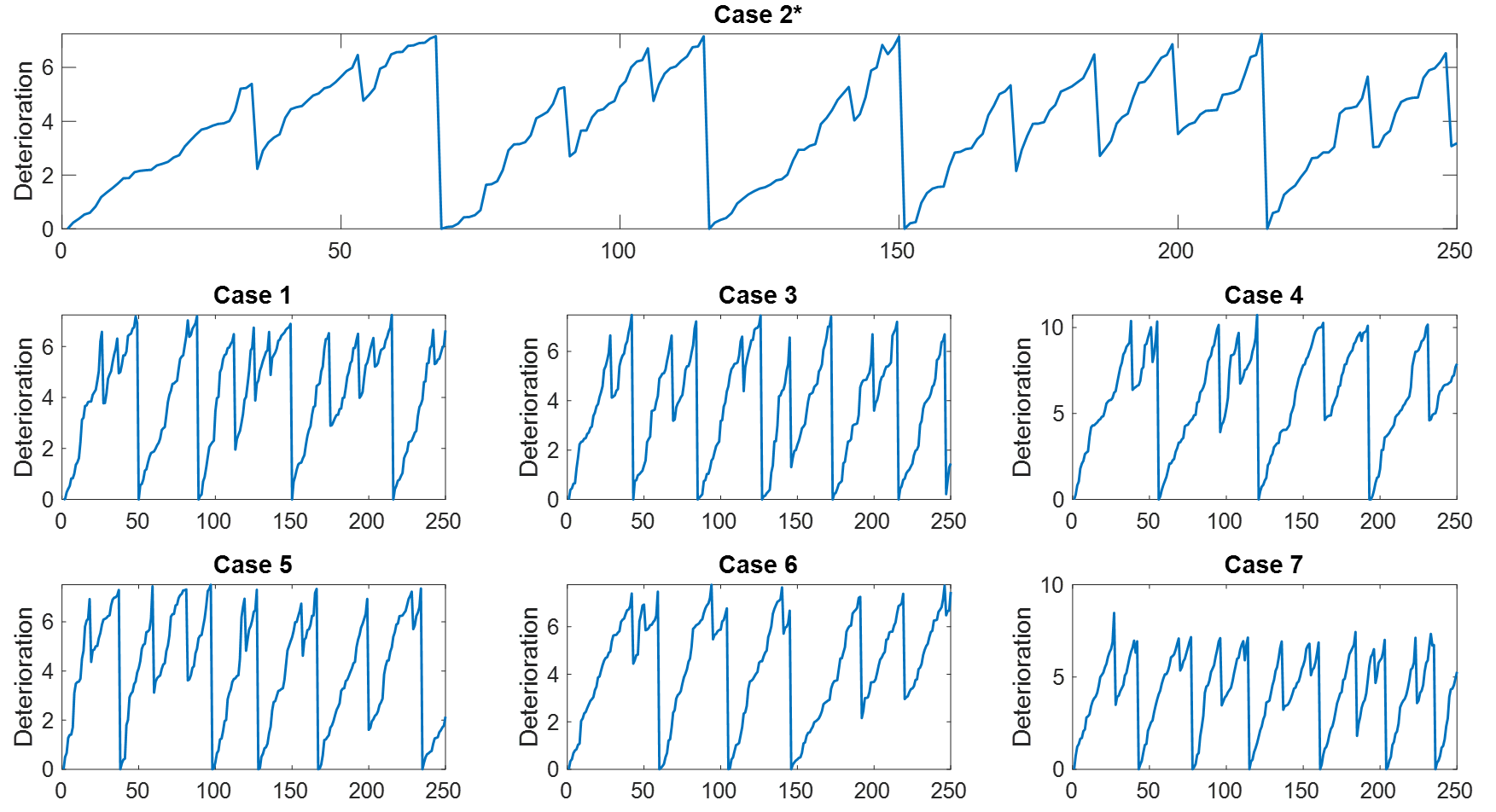}
		\caption{RL-based maintenance for all case studies}\label{fig:RLBM}
	\end{figure}

	The proposed agent has been trained for the scenarios in Table \ref{tab:Tab1}, providing specific maintenance policies for each case. \jp{Figure \ref{fig:RLBM}} shows the system deterioration for the baseline case (Case 2*) and the rest of the cases using the policies proposed by the agent. Each scenario comprises a maintenance period of $1000$ inspections, but only $250$ inspections are shown for the sake of clear visualization. 
	
\begin{figure}[!h]
		\centering
		\includegraphics[trim={0 2cm 0 1cm}, width=12cm]{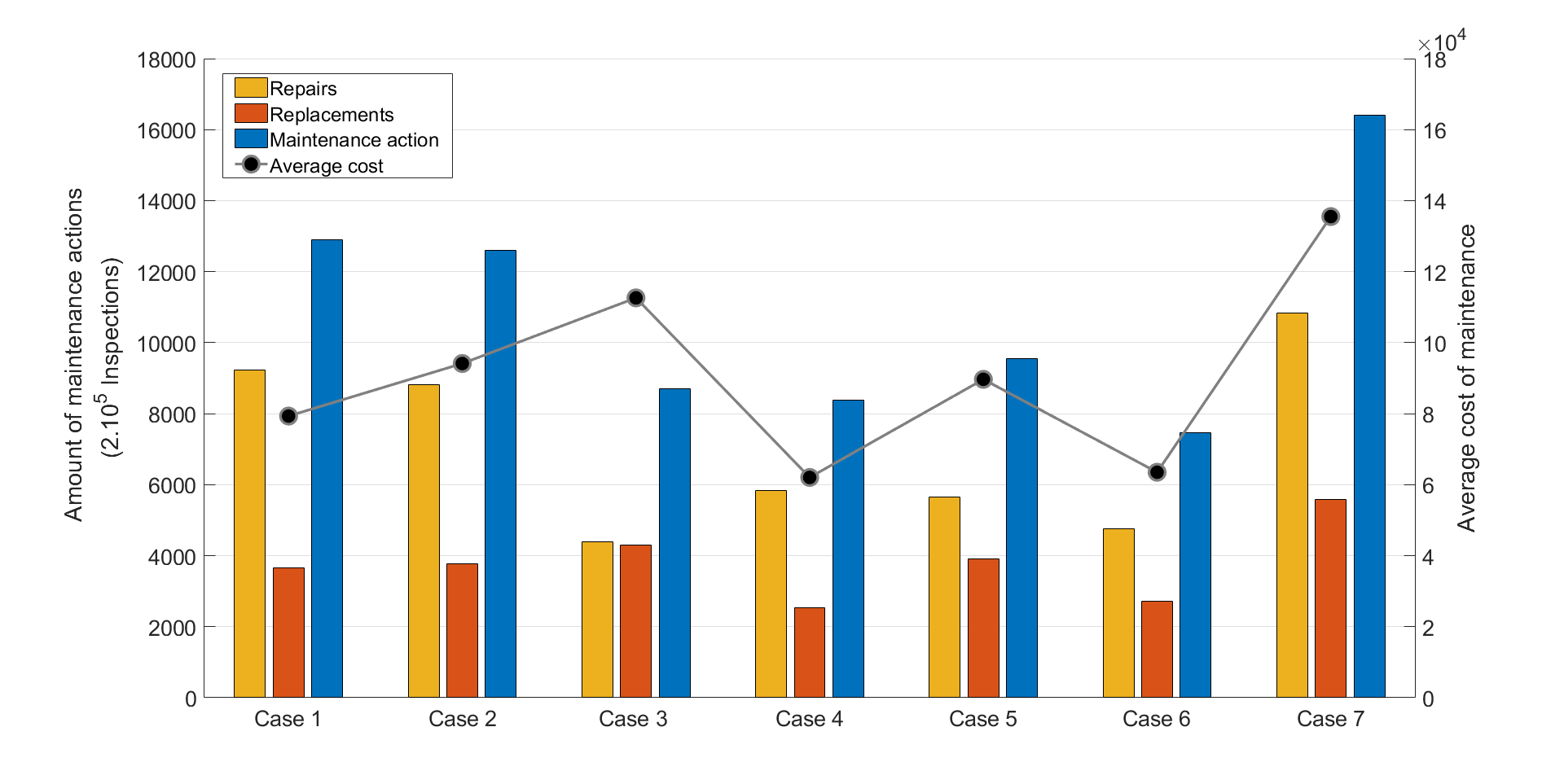}
		\caption{Amount of maintenance actions}\label{fig:AMA}
	\end{figure}

\begin{figure}[!h]
		\centering
		\includegraphics[trim={0 1.5cm 0 1cm}, width=13cm]{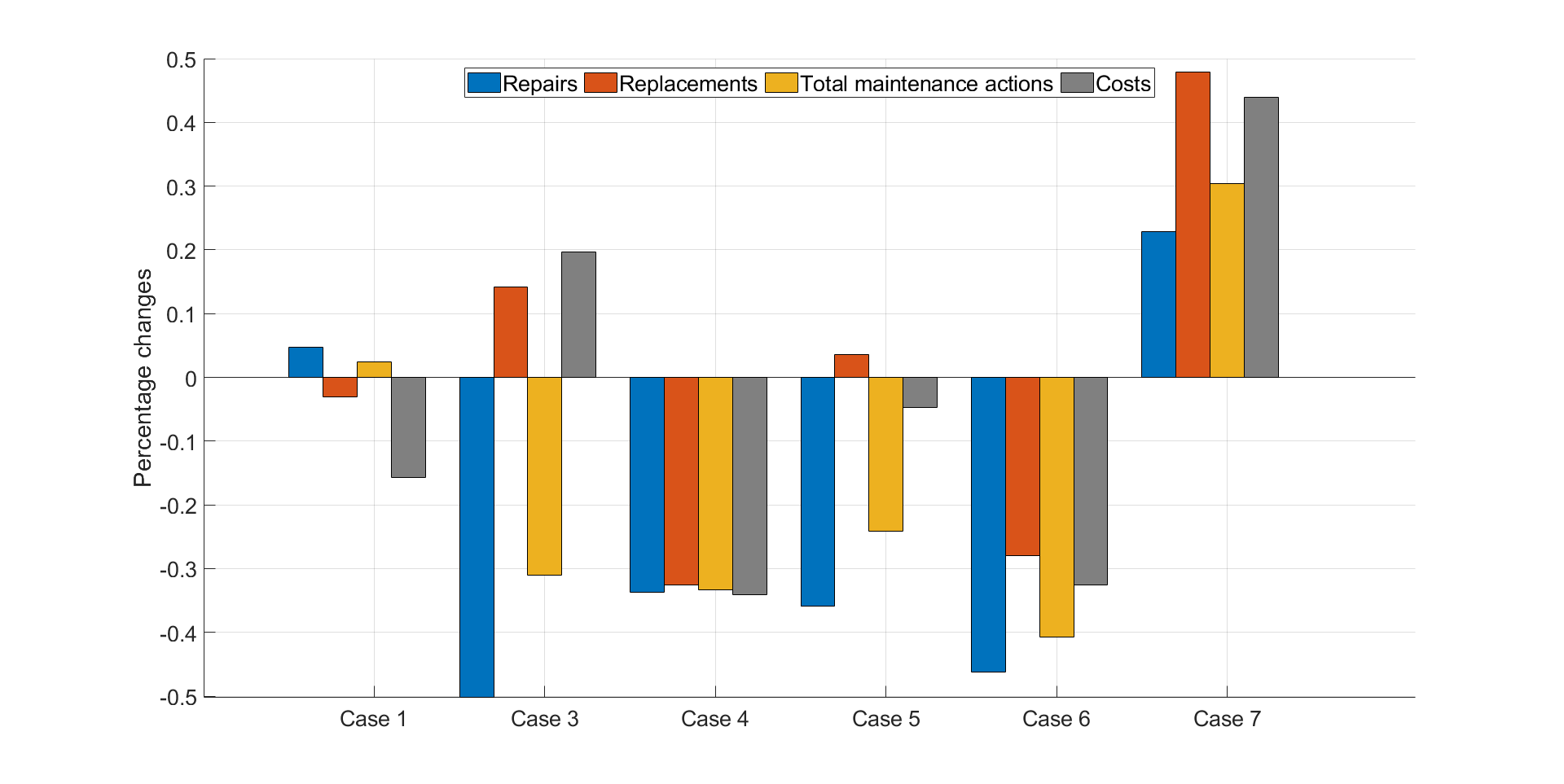}
		\caption{Percentage changes in the number of maintenance actions and costs with respect to Case 2*}\label{fig:PCNMA}
	\end{figure}

	\jp{Figure \ref{fig:RLBM}} allows us to conduct an initial analysis of how the agent performs in each scenario. For instance, it is clear that more repairs are carried out in Case 1, unlike Case 3, where fewer repairs are conducted. Therefore, the policy in Case 1 provides longer renewal cycles. It can be observed that in Case 6, due to slower degradation, the system is pushed closer to the failure threshold before a repair is executed, meaning that the agent allows greater risks of degradation exceeding such threshold. These and other observations are quantitatively analyzed in \jp{Figure \ref{fig:AMA}} and \jp{Figure \ref{fig:PCNMA}} . These figures are based on complete periods of $1000$ inspections and $200$ Monte Carlo iterations. Therefore, the results are based on a total of $200,000$ inspections.
	
	\jp{Figure \ref{fig:AMA}} shows the total amount of maintenance actions that the RL agent proposes for each case. Moreover, the black line represents the average costs of maintenance for the Monte Carlo iterations. 
	
	\jp{Figure \ref{fig:PCNMA}} shows the percentage changes for each type of maintenance action with respect to Case 2*. 
	
	\jp{Figure \ref{fig:AMA}} and \jp{Figure \ref{fig:PCNMA}} provide some important information regarding the maintenance policy generated for each case study. With respect to Case 2* we observe that:
	\begin{itemize}
		\item Repairs are cheaper in Case 1 and therefore the policy increases $4.7\%$ the number of repairs and decreases $3.7\%$ the number of replacements. These variations are rather small, but they lead to a reduction of $15.7\%$ in the average costs. 
		\item In Case 3, repairs are more expensive, and the policy drastically reduces the number of repairs by $50\%$. This forces the agent to carry out $14.1\%$ more replacements. It is worth mentioning that although there is a reduction of more than $30\%$ in the total number maintenance actions, the average maintenance costs increase significantly. 
		\item In Case 4, the failure threshold is higher and therefore the number of maintenance actions and the average costs in the same time period are reduced. However, the ratio between repairs and replacements remains similar since the costs of maintenance actions have not changed. Therefore, a change in the failure threshold will affect the maintenance policy in terms of "when" but not "which" maintenance actions should be performed. 
		\item Case 5 presents lower costs of corrective replacements through a reduction of downtime costs. We observe that the agent assumes more risk to perform a preventive replacement since, upon surpassing the maximum threshold $L$, the penalization is less significant. Therefore, the number of replacements increases by $3.6\%$, leading to a significant reduction of repairs ($35.9\%$), which causes a slight reduction in costs. 
		\item Case 6 considers that the degradation process is slower. Therefore, both the total number of maintenance actions and the mean costs of maintenance decrease. In general, this scenario is favorable in every way. It is clear that if the system degradation is slower, the number of maintenance activities of any type is reduced, and consequently the costs also decrease. 
		\item Case 7 involves a variation in the time between inspections. If the period between inspections is longer, it is more likely that the maximum threshold $L$ will be reached and therefore more corrective actions must be taken. In addition, preventive actions are proposed at lower deterioration states in order to avoid the risk of surpassing the threshold. It is worth mentioning that we are not considering costs of inspections, which would lead to savings in this case study. 
	\end{itemize}
	
	In general, we can observe how the RL agent is able to learn from each case study and adapt the maintenance policy to the specificities of each case. Table \ref{tab:summ} shows some results (mean and standard deviation) obtained from the analysis.

	\begin{table}[htbp]
		\centering
		\caption{Summary of case-study results}
		\footnotesize
			\begin{tabular}{ccccccccc}
				\cline{2-9}
				& \multicolumn{2}{p{1.7cm}}{N. of Repairs $(N_P)$} & \multicolumn{2}{p{2.2cm}}{Number of Preventive Replacements $(N_{PR})$} & \multicolumn{2}{p{2.2cm}}{Number of Corrective Replacements $(N_{CR})$} & \multicolumn{2}{p{2cm}}{Renewal Cycles Duration $(S)$} \\
				\cline{2-9}
				& Mean & sd & Mean  & sd & Mean  & sd & Mean & sd\\
				\cline{1-9}    
				\textbf{Case 1} & 46.19 & 3.17  & 17.99 & 1.22  & 0.28  & 0.53  & 53.33 & 3.25 \\
			\textbf{Case 2} & 44.12 & 1.87  & 18.54 & 1.14  & 0.31  & 0.55  & 51.70 & 2.87 \\
				\textbf{Case 3} & 21.95 & 0.79  & 20.71 & 1.15  & 0.80  & 0.88  & 45.37 & 1.57 \\
				\textbf{Case 4} & 29.23 & 1.46  & 12.72 & 0.98  & 0.00  & 0.00  & 76.26 & 5.72 \\
				\textbf{Case 5} & 28.27 & 1.57  & 18.43 & 1.48  & 1.10  & 1.05  & 49.99 & 2.89 \\
				\textbf{Case 6} & 23.75 & 1.23  & 13.25 & 1.04  & 0.32  & 0.56  & 71.36 & 4.96 \\
				\textbf{Case 7} & 54.22 & 2.05  & 27.55 & 1.42  & 0.32  & 0.56  & 35.27 & 1.67 \\
			\end{tabular}%
			%}
		\label{tab:summ}%
	\end{table}%
	
	Using these statistical results, confidence intervals are used to estimate the expected values required to calculate long-run cost rates. By performing Anderson-Darling and Kolmogorov-Smirnov tests, we have verified that all the collected parameters are normally distributed in the Monte Carlo iterations. Therefore, the $95\%$ confidence intervals for the long-run cost rate are given in Table \ref{tab:confInt} . 

	\begin{table}[htbp]
		\centering
		\caption{Relevant confidence intervals}
		\resizebox{\linewidth}{!}{
			\begin{tabular}{ccccccccccc}
				\cline{2-11}
				& \multicolumn{2}{c}{Interval for $N_P$} & \multicolumn{2}{c}{Interval for $N_{PR}$} & \multicolumn{2}{c}{Interval for $N_{CR}$} & \multicolumn{2}{c}{Interval for $(S)$} & \multicolumn{2}{c}{Long Run Cost Rate} \\
				\cline{2-11}  & Lower  & Upper  & Lower  & Upper  & Lower  & Upper  & Lower  & Upper  & Lower & Upper \\
				\cline{1-11}    \textbf{Case 1} & 45.75 & 46.63 & 17.82 & 18.16 & 0.21  & 0.35  & 52.88 & 53.78 & 1436  & 1503 \\
				 \textbf{Case 2} & 43.86 & 44.38 & 18.38 & 18.70 & 0.23  & 0.39  & 51.30 & 52.10 & 1764  & 1836 \\
				\textbf{Case 3} & 21.84 & 22.06 & 20.55 & 20.87 & 0.68  & 0.92  & 45.15 & 45.59 & 2378  & 2462 \\
				 \textbf{Case 4} & 29.03 & 29.43 & 12.58 & 12.86 & 0.00     & 0.00     & 75.47 & 77.05 & 797   & 830 \\
				\textbf{Case 5} & 28.05 & 28.49 & 18.22 & 18.64 & 0.95  & 1.25  & 49.59 & 50.39 & 1675  & 1760 \\
			\textbf{Case 6} & 23.58 & 23.92 & 13.11 & 13.39 & 0.24  & 0.40  & 70.67 & 72.05 & 851   & 897 \\
				\textbf{Case 7} & 53.94 & 54.50 & 27.35 & 27.75 & 0.24  & 0.40  & 35.04 & 35.50 & 3645  & 3767 \\
			\end{tabular}
		}
		\label{tab:confInt}%
	\end{table}
	
\begin{figure}[!h]
		\centering
		\includegraphics[trim={0 2.5cm 0 1cm}, width=12cm]{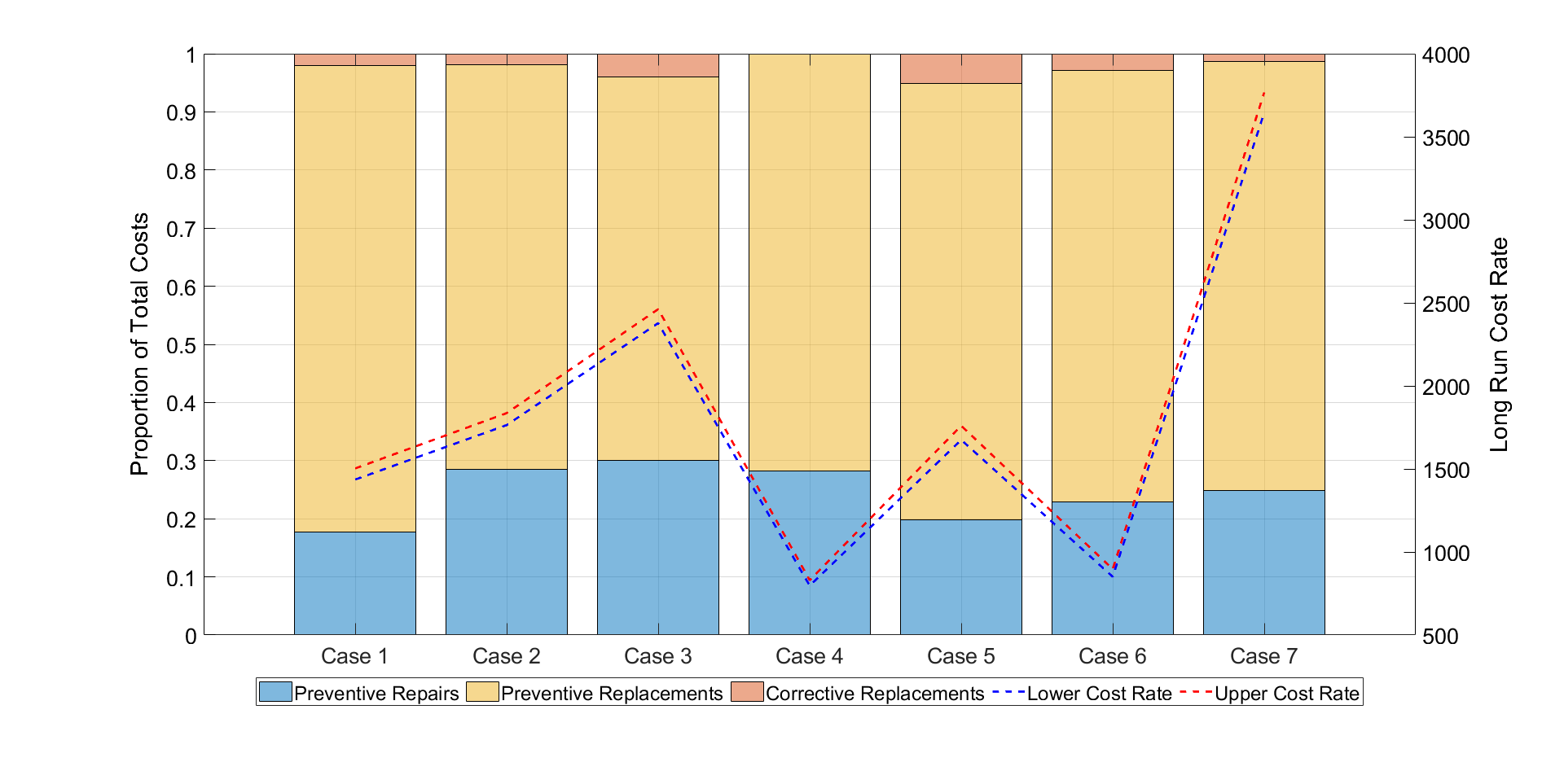}
		\caption{Long-run cost rates and distribution of costs for Cases 1–7}\label{fig:LRCR}
	\end{figure}

	By considering the central point of each interval, we calculate which part of the long-run cost rate corresponds to each type of maintenance action, obtaining the results presented in \jp{Figure \ref{fig:LRCR}} . 
	
	The long-run cost rate has a very similar shape to the average costs shown in \jp{Figure \ref{fig:AMA}} . It is worth mentioning that average cost corresponds to a complete period of $1000$ inspections and $200$ iterations; however, the long-run cost rates correspond to the average costs during a single renewal cycle, i.e., from a replacement to the moment just before the next replacement. In general, we observe that, independently to the parameters configuration, around $70\%$, $27\%$, and $3\%$ of long-run cost is due to preventive replacements, repairs, and corrective replacements, respectively. 
	
	\jp{It is worth mentioning that maintenance plays a crucial role in both availability and productivity. In our model, maintenance activities are assumed to be instantaneous and therefore the system is available unless deterioration reaches the failure threshold. An important factor which would affect the system availability is the duration of maintenance activities, however, since this model is not described for a specific system, it is not convenient to provide such durations. However, the impact of maintenance on availability and productivity is partially captured by the parameter $C_{down}$. Consequently, the product of the total number of corrective replacements $(N_CR )$ and $C_{down}$ can be used as relative measure of unavailability to compare between the different cases of study as shown in Table \ref{tab:IAva} . }

	\begin{table}[htbp]
		\centering
		\caption{\jp{Impact on Availability}}
		\small
		\begin{tabular}{ccccc}
			& $E[N_{CR}]$ & $C_{down}$ & $E[N_{CR}] \cdot C_{down}$ & Availability Ranking \\
			\hline
			 \textbf{Case 1} & 0.28  & 2000  & 560   & $4^{th}$ \\
			\textbf{Case 2} & 0.31  & 2000  & 620   & $3^{rd}$ \\
		\textbf{Case 3} & 0.80   & 2000  & 1600  & $7^{th}$ \\
			\textbf{Case 4} & 0.00     & 2000  & 0     & $1^{st}$ \\
		\textbf{Case 5} & 1.10   & 500   & 550   & $2^{nd}$ \\
			\textbf{Case 6} & 0.32  & 2000  & 640   & $5^{th}$ \\
		 \textbf{Case 7} & 0.32  & 2000  & 640   & $6^{th}$ \\
		\end{tabular}
		\label{tab:IAva}
	\end{table}

	\jp{As can be observed in Table \ref{tab:IAva} , Case 4 presents the highest availability since it has the highest failure threshold. Cases 6 and 7 presents the result for the product $E[N_{CR}] \cdot C_{down}$ , however, Case 6 has been positioned before since the total number of maintenance actions is three times less than in Case 7. Finally, in Case 3, the elevated cost of preventive activities impacts negatively on the availability.} 
	
\subsection{Comparison to conventional maintenance policies}
	To determine the validity of the proposed procedure, we performed a comparison between the performance of the proposed RL agent and other conventional CBM strategies for Case 2*. \jp{In this paper, the RL Agent has the goal of improving maintenance from an economic perspective, i.e. the only objective is to minimize maintenance long run cost rate considering that when deterioration is above the failure threshold, a corrective maintenance action must be immediately done. Therefore, our maintenance model is defined in such way that failures do not cause safety problems or environmental risks but only economic losses. Additionally, system reliability is not considered to be an objective function in this paper. These are the reasons why some maintenance strategies such as risk-based or reliability centered maintenance are not included in this comparison.}  
	
	Similarly to Andriotis and Papakonstantinou \citep{andriotis2021deep}, we consider the following policies:	
	\begin{itemize}
		\item \textit{Fail Replacement (FR) policy}: Only corrective replacements are permitted. In this policy a corrective replacement is performed when the deterioration of the system is above the failure threshold $L$. 
		\item \textit{Age-based Periodic Maintenance policy}: This policy assumes that repairs and replacements are done periodically. Therefore, two important parameters must be defined: the time period between consecutive repairs and the time period between consecutive replacements. In order to compare with the proposed RL based policy, both time periods have been optimized numerically with Monte Carlo iterations.
		\item \textit{Threshold-based Maintenance (TBM) policy}: Maintenance actions are taken depending on the current state of deterioration of the system at the inspection time. Two thresholds are set and optimized, i.e., a preventive threshold, to determine when a preventive replacement is performed, and a corrective threshold, to define when a corrective replacement is required. In order to compare with the proposed RL based policy, both thresholds have been previously optimized numerically. 
		\item \jp{\textit{Age and Threshold-based Maintenance (ATBM) policy}: Maintenance actions are taken depending on both the current state of deterioration of the system and a certain time period between consecutive repairs and replacements. Four parameters have been considered in this strategy, i.e. two thresholds to determine if a preventive action or a corrective action must be done and two time periods to determine when a repair and replacement must be done. In order to compare with the proposed RL based policy, the four parameters (thresholds and time periods) have been previously optimized numerically.}
	\end{itemize}	

\begin{figure}[!h]
		\centering
		\includegraphics[trim={0 2cm 0 1cm}, width=12cm]{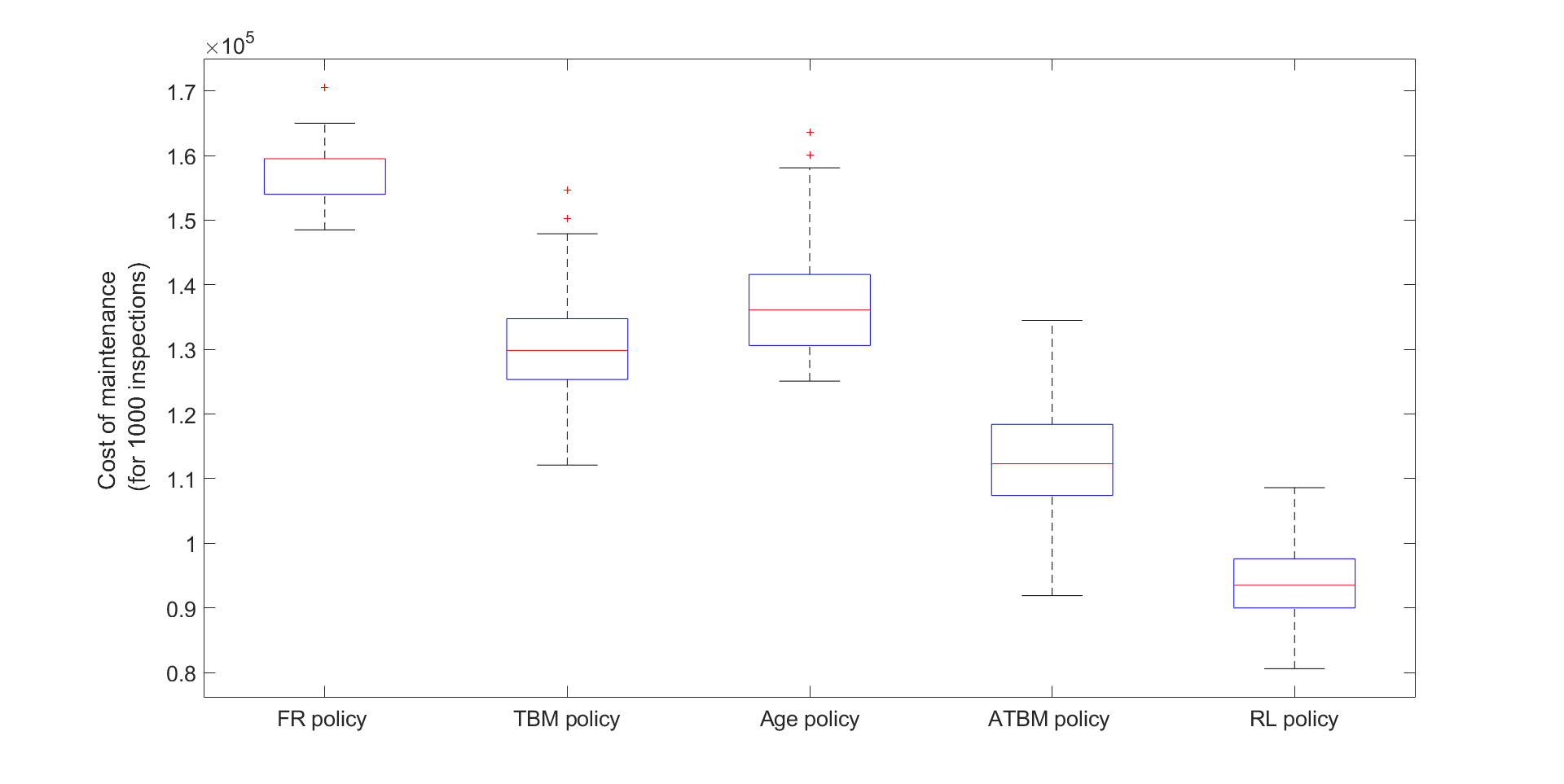}
		\caption{Comparison with other policies}\label{fig:ComPoli}
	\end{figure}

	\jp{Figure \ref{fig:ComPoli}} shows the costs of maintenance for each policy in a total of $200$ iterations.
	
	\jp{Figure \ref{fig:ComPoli}} shows that the agent is able to reduce the long-run cost rate by around $41\%$, $28\%$, \jp{$31\%$ and $17\%$} compared with the FR policy, TBM policy, \jp{Age policy, and ATBM policy,} respectively. Therefore, the RL Agent proposed in this paper clearly outperforms other conventional maintenance policies.
	
\section{Conclusions}\label{S7}
	This study successfully developed a homogeneous gamma degradation model whose maintenance framework is based on periodic and perfect inspections, i.e., inspections reveal the real degradation level of the system. Two type of maintenance actions were considered: repairs or replacements. These actions are categorized as either corrective or preventive depending on the state of system at the time the action is carried out. 
	
	A model has been proposed wherein repair actions enhance the degradation following a probability distribution, representing imperfect maintenance subject to uncontrollable conditions. A novel feature of this model is that each repair action negatively affects the effectiveness of the subsequent repair by affecting the parameters of the probability distribution. 
	
	To optimize maintenance tasks, we implemented an RL agent with a DDQN structure, demonstrating its capability to decide when and what maintenance activities are advisable in different scenarios. One of the main advantages of this approach is that there is no requirement to define a preventive threshold. The RL-based agent discerns the ideal timing for executing corrective or preventive maintenance autonomously. In addition, this RL architecture was demonstrated to be highly effective when facing large or continuous state space. Another novelty if this study is the capacity of our RL agent to make decisions without discretizing the degradation variable. 
	
	Additionally, an analysis has been conducted to understand how each parameter influences the long-term maintenance costs based on the adopted policy. This study has demonstrated that the RL agent is able to create flexible policies adapted to changing environments. 
	
	Finally, the model was validated, revealing that our agent significantly improves long-term costs compared to \jp{other maintenance} policies.

\section*{\jp{Acronyms}}

	\begin{description}
		\item[ADAM:] Adaptive Learning Rates
		\item[AGAN:] As Good as New
		\item[API-CBM:] Age-Periodic Inspections with Condition-Based Maintenance 
		\item[APM:] Age-Periodic Maintenance
		\item[\jp{ATBM:}] \jp{Age and Threshold-based Maintenance} 
		\item[CBM:] Condition Based Maintenance 
		\item[CM:] Corrective Maintenane 
		\item[DDQN:] Double Deep Q-Network 
		\item[DDMAC:] Deep Centralized Multi-Agent Actor Critic
		\item[DQN:] Deep Q-Network 
		\item[FR:] Fail Replacement
		\item[GPRL:] Gaussian Process with Reinforcement Learning 
		\item[MDP:] Markov Decision Process
		\item[O\&M:] Operation and Maintenance
		\item[PdM:] Predictive Maintenance
		\item[PM:] Preventive Maintenance
		\item[PPO:] Proximal Policy Optimization
		\item[RBI-CBM:] Risk-Based Inspections with Condition-Based Maintenance
		\item[RL:] Reinforcement Learning
		\item[RUL:] Remaining Useful Life
		\item[SDP:] Stochastic Deterioration Process
		\item[TPI-CBM:] Time-Periodic Inspections with Condition-Based Maintenance
		\item[TRPO:] Trust Region Policy Optimization
	\end{description}

\section*{Acknowledgement}	

The work reported herewith has been financially supported by the Spanish Ministerio de Ciencia, Innovación y Universidades, under Research Grant FOWFAM project with reference: PID2022-140477OA-I00.

\bibliographystyle{elsarticle-num}
\bibliography{RLG_Bibli.bib}

\end{document}